\PassOptionsToPackage{table,dvipsnames}{xcolor}
\documentclass[10pt,twocolumn,letterpaper]{article}
\usepackage[pagenumbers]{iccv}

\usepackage[accsupp]{axessibility}
\usepackage{multirow}
\usepackage[misc]{ifsym}
\usepackage{siunitx}

\makeatletter
\renewcommand*{\@fnsymbol}[1]{%
  \ifcase#1
    \or $\dagger$
    \or $\ddagger$
    \else \@ctrerr
  \fi
}
\makeatother

\definecolor{iccvblue}{rgb}{0.21,0.49,0.74}
\usepackage[pagebackref,breaklinks,colorlinks,allcolors=iccvblue]{hyperref}

\newcommand{\OURS}{DepR}

\title{\OURS{}: Depth Guided Single-view Scene Reconstruction with Instance-level Diffusion}

\author{
Qingcheng Zhao$^{*,1,}$\thanks{Work done during internship at UC San Diego.}
\quad Xiang Zhang$^{*,}$\textsuperscript{\Letter}$^{,2}$
\quad Haiyang Xu$^2$
\quad Zeyuan Chen$^2$ \\
\quad Jianwen Xie$^3$
\quad Yuan Gao$^4$
\quad Zhuowen Tu$^2$ \\
$^1$ShanghaiTech University \quad $^2$UC San Diego \quad $^3$Lambda, Inc. \quad $^4${Stanford University} \\
{\small $^*$equal contribution \quad \textsuperscript{\Letter}corresponding author}
}

\begin{document}

\maketitle

\captionsetup{belowskip=-5pt,aboveskip=2pt}

\begin{abstract}

We propose \OURS{}, a depth-guided single-view scene reconstruction framework that integrates instance-level diffusion within a compositional paradigm. Instead of reconstructing the entire scene holistically, \OURS{} generates individual objects and subsequently composes them into a coherent 3D layout. Unlike previous methods that use depth solely for object layout estimation during inference and therefore fail to fully exploit its rich geometric information, \OURS{} leverages depth throughout both training and inference. Specifically, we introduce depth-guided conditioning to effectively encode shape priors into diffusion models. During inference, depth further guides DDIM sampling and layout optimization, enhancing alignment between the reconstruction and the input image. Despite being trained on limited synthetic data, \OURS{} achieves state-of-the-art performance and demonstrates strong generalization in single-view scene reconstruction, as shown through evaluations on both synthetic and real-world datasets.

\end{abstract}
\section{Introduction}
\label{sec:intro}

Single-view scene reconstruction is a long-standing challenge in computer vision, requiring the estimation of 3D object shapes and scene structure from a single 2D image. The task is inherently difficult due to the limited spatial cues available in a single-view observation, making it heavily reliant on prior knowledge to infer plausible 3D geometry.

Previous approaches performing holistic scene reconstruction~\cite{dahnert2021panoptic,chu2023buol,zhang2023uni} have successfully leveraged depth to back-project features into 3D volumes, followed by an encoder-decoder architecture to reconstruct scene-level geometries. Depth plays a crucial role in these methods, facilitating effective shape reconstruction. However, the limited resolution and scarcity of scene-level training data hinder their ability to generalize effectively to real images.

Recently, with the advancement of single-view object reconstruction/generation models \cite{hong2023lrm,liu2023zero,liu2023one,liu2024one,jun2023shap} alongside the emergence of large-scale object datasets \cite{deitke2023objaverse,deitke2023objaversexl}, stage-wise compositional scene generation has gained increasing popularity. Notably, methods such as DeepPriorAssembly~\cite{zhou2024zero} and Gen3DSR~\cite{dogaru2024generalizable} utilize off-the-shelf 2D segmentation models to extract individual objects from images, then pass the masked images into a pre-trained image-to-shape or single-image-to-multi-view generative model. A monocular depth estimator is subsequently used to optimize object poses, ensuring proper spatial arrangement.

While these methods benefit from the generalizability of pre-trained object reconstruction models, they have two key limitations: (1) they overlook the strong shape priors provided by depth; and (2) they rely on reconstruction models pre-trained on complete object images, neglecting occlusions and scene-level context, which can cause inconsistencies between the reconstruction and the input image.

\begin{figure}[t]
    \centering
    \includegraphics[width=\linewidth]{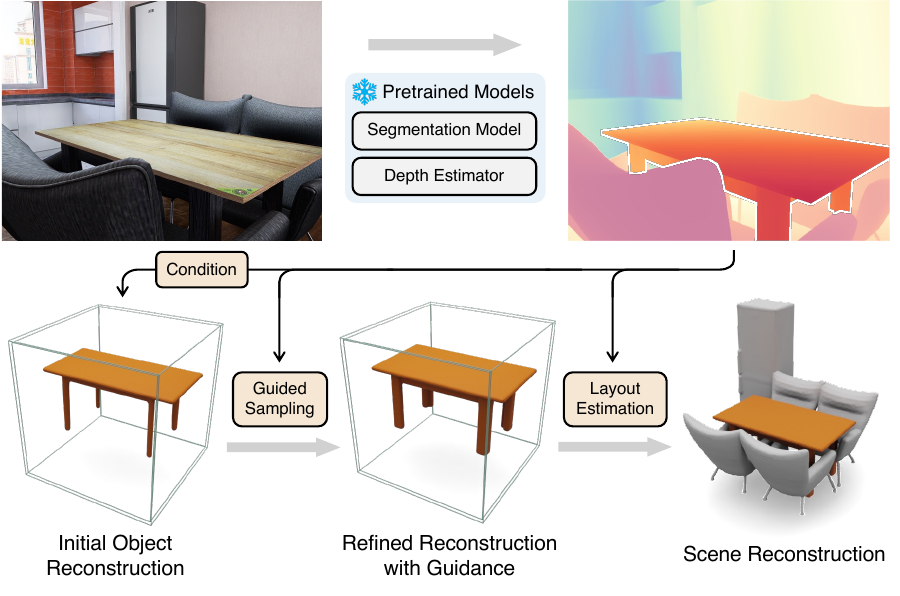}
    \caption{Illustration of \OURS{}. Given a single 2D input image, we leverage depth predicted by a pretrained model to condition object reconstruction and guide sampling, followed by layout estimation to assemble the reconstructed instances into a coherent scene.}
    \label{fig:teaser}
\end{figure}

To overcome these limitations, we propose \OURS{}, a depth-guided single-view scene reconstruction pipeline following a compositional generation paradigm. Our method begins with off-the-shelf pre-trained segmentation and monocular depth estimation models, but rather than using depth solely for object pose alignment, we incorporate depth priors directly into the object reconstruction network via depth-guided conditioning. To handle partial observations caused by object occlusions, we introduce local-global attention, which leverages scene-level context to enhance the conditioning signal. The depth guidance enables our image-conditioned shape generative model to generalize effectively to real images, even when trained on limited synthetic data. At inference time, we further utilize depth for guided DDIM sampling and layout estimation, ensuring accurate alignment between the reconstruction and the input.

Our contributions can be summarized as follows:

\begin{itemize}
    \item We introduce a novel framework for single-view scene reconstruction that effectively leverages a pre-trained monocular depth estimator to facilitate both object reconstruction and layout estimation.
    \item We formulate object reconstruction as an image-conditioned generation task, incorporating \textit{depth-guided conditioning} and \textit{local-global attention} to leverage depth priors while integrating global scene context.
    \item To improve alignment between the generated reconstructions and input images, we introduce depth-guided sampling, which enhances geometric consistency. Experiments demonstrate that \OURS{} produces accurate reconstructions that closely align with input images.
\end{itemize}

\section{Related Work}
\label{sec:related}

\subsection{Single-view Object Reconstruction}

Recent advancements in large-scale datasets \cite{deitke2023objaverse,deitke2023objaversexl} have significantly progressed 3D object reconstruction/generation. Several approaches \cite{liu2023zero,liu2023one,liu2024one,liu2023syncdreamer,long2024wonder3d,huang2024epidiff,wu2024unique3d,wang2024crm} leverage priors from image \cite{rombach2022high,podellsdxl} or video generative models \cite{blattmann2023stable} to synthesize multi-view consistent images from a single view. These generated views are processed by feed-forward reconstruction frameworks \cite{tang2024lgm,xu2024grm,voleti2024sv3d} to construct an underlying 3D representation, such as Gaussian splatting \cite{kerbl20233d} or neural radiance fields (NeRFs) \cite{mildenhall2021nerf}. Other methods directly generate 3D representations, such as triplanes \cite{shue20233d}, using large Transformer-based architectures \cite{hong2023lrm,zou2024triplane,xu2024instantmesh,zhang2024clay,xu2024bayesian} or diffusion models \cite{wu2024blockfusion,yan2024frankenstein,wudirect3d,zhao2023michelangelo}, eliminating the need for intermediate multi-view image synthesis. 

\subsection{Single-view Scene Reconstruction}

Reconstructing a 3D scene from a single image is a challenging problem due to the inherent depth ambiguity and occlusions. Existing approaches can be broadly categorized into two paradigms based on how they handle scene composition and object representation: holistic scene reconstruction and compositional scene reconstruction.

\paragraph{Holistic scene reconstruction.} Holistic scene reconstruction methods \cite{dahnert2021panoptic,zhang2023uni,chu2023buol} reconstruct the entire scene as a whole, grouping individual instances based on predicted instance labels. Monocular depth estimation \cite{hu2019revisiting,gao2022panopticdepth} is often employed as a strong shape prior, facilitating the back-projection of 2D features into a 3D volume. This is typically followed by an encoder-decoder architecture \cite{ronneberger2015u} to recover the scene geometry. While these approaches naturally capture object poses and overall scene layout, they often suffer from limited resolution and availability of high-quality training data, leading to poor generalization to real-world images and reduced reconstruction fidelity.

\vspace{-1em}

\paragraph{Compositional scene reconstruction.} Compositional scene reconstruction approaches decompose a scene into individual objects, reconstruct them separately, and subsequently estimate the scene layout and object poses to reintegrate them into the scene space.  Earlier feed-forward methods utilize either explicit mesh representations \cite{huang2018holistic,nie2020total3dunderstanding,liu2022towards,gkioxari2022learning}, or implicit signed distance functions (SDFs) \cite{zhang2021holistic,liu2022towards,chen2024single}, trained end-to-end with 3D supervision. While these methods are relatively efficient to train and infer, they exhibit limited generalizability to out-of-domain images, particularly in complex real-world scenarios with variations not seen during training.

Another line of research \cite{izadinia2017im2cad,kuo2020mask2cad,kuo2021patch2cad,gumeli2022roca,wu2022casa,langer2022sparc,yan2023psdr,gao2024diffcad} focuses on retrieval-based approaches, where 3D models are selected from a database and aligned to the scene. While these methods benefit from high-quality pre-existing 3D assets, their reconstruction accuracy is inherently constrained by the diversity and completeness of the database. Additionally, insufficient visual cues may lead to misalignment between retrieved models and the actual scene content.

More recently, generation-based methods \cite{chen2024comboverse,dogaru2024generalizable,han2024reparo,zhou2024zero} have emerged as a promising direction, leveraging large-scale generative models for both images \cite{kirillov2023segment,ren2024grounded,liu2024grounding,nichol2021glide,rombach2022high,podellsdxl,ramesh2022hierarchical,saharia2022photorealistic} and 3D shapes \cite{eftekhar2021omnidata,jun2023shap,zhao2023michelangelo,liu2023zero,zhang2024clay}. These approaches typically follow a pipeline where objects are first segmented using pre-trained segmentation models \cite{kirillov2023segment,ren2024grounded}, then reconstructed from their segmented images using large-scale object reconstruction models \cite{liu2023zero,jun2023shap}, and finally refined through pose optimization to ensure spatial coherence within the scene. While the use of large-scale pre-trained object reconstruction models enhances reconstruction quality, these methods lack holistic scene context. Furthermore, since object reconstruction models are generally trained on complete object images, they struggle with occluded or incomplete segmented objects, necessitating amodal completion to infer missing regions.

\section{Method}
\label{sec:method}

\begin{figure*}[htbp]
  \centering
  \includegraphics[width=\textwidth,trim=0 0.5em 0 0,clip]{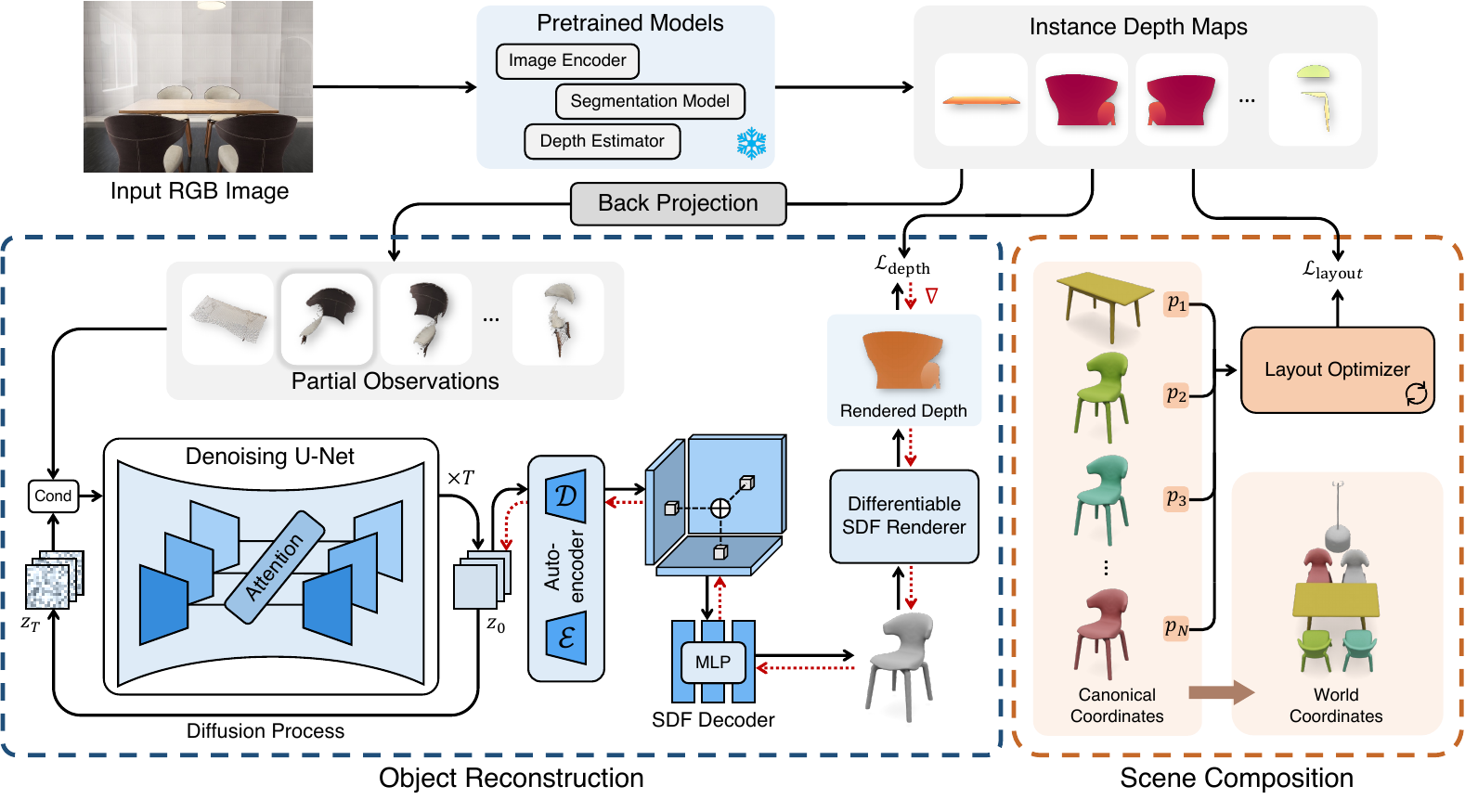}
  \caption{Overview of our method. Depth is utilized in three key stages: 1) to back-project features to condition the latent tri-plane diffusion model to generate complete 3D shapes; 2) to guide the diffusion sampling process via gradients from a depth loss $\mathcal{L}_\mathrm{depth}$; and 3) to optimize object poses $\{p_i\}_{i=1}^N$ via $\mathcal{L}_\mathrm{layout}$ for accurate scene composition.}
  \label{fig:pipeline}
\end{figure*}

Our method, illustrated in \cref{fig:pipeline}, reconstructs a complete 3D scene from a single RGB image using a multi-stage pipeline that leverages conditional generation. We begin by formulating the problem addressed by \OURS{} in \cref{sec:problem_formulation}, followed by a description of our object-level reconstruction approach in \cref{sec:obj_recon}. Finally, \cref{sec:scene_comp} details our layout optimization strategy, which composes the reconstructed objects back into a coherent scene.

\subsection{Problem Formulation}\label{sec:problem_formulation}

Single-view scene reconstruction pipelines aim to recover the geometry of a scene from a single RGB image. Following the setting of InstPIFu~\cite{liu2022towards} and DeepPriorAssembly~\cite{zhou2024zero}, we focus exclusively on object instances (\eg sofas, chairs, \etc), while excluding background elements such as walls. Unlike holistic reconstruction approaches such as Uni-3D~\cite{zhang2023uni} and PanoRe~\cite{dahnert2021panoptic}, which treat the entire scene as a single mesh, our stage-wise pipeline reconstructs each object instance individually.

Given an RGB image $I \in \mathbb{R}^{H \times W \times 3}$, we leverage off-the-shelf models to obtain the depth map $D$, instance segmentation masks $\mathcal{M} = \{M_i\}_{i=1}^N$, and image features $F$. For each object instance, we extract its depth $D_i = D \odot M_i$ and corresponding image features $F_i = F \odot M_i$, where $\odot$ represents element-wise multiplication. These serve as inputs to the object reconstruction model $F_{\mathrm{recon}}$:
\begin{equation}
    O_i = F_{\mathrm{recon}} (F_i, D_i, F)
\end{equation}
where $O_i$ denotes the reconstructed object shape in its canonical coordinate frame. Due to self-occlusion, the depth $D_i$ captures only the visible front surface of the object. Furthermore, inter-object occlusion often results in incomplete segmentation masks $M_i$. These limitations make it difficult for feedforward methods such as InstPIFu to accurately reconstruct full object geometry. To address this, we build $F_{\mathrm{recon}}$ as a generative model conditioned on partial observations.

To reassemble the full scene, a layout optimizer $F_\mathrm{layout}$ can be employed to estimate the pose $p_i$ of each object:
\begin{equation}
    p_i = F_{\mathrm{layout}} (O_i, D_i)
\end{equation}

The final scene reconstruction is given by the set $\{ O_i, p_i \}_{i=1}^N$, which specifies the geometry and spatial configuration of all object instances.

\subsection{Object Reconstruction from Partial Observations}\label{sec:obj_recon}

In this section, we introduce our object reconstruction model. Inspired by works such as BlockFusion~\cite{wu2024blockfusion} and Frankenstein~\cite{yan2024frankenstein}, we adopt the latent tri-plane representation to enable compact and efficient encoding of 3D geometry. We begin with tri-plane fitting, followed by a variational autoencoder (VAE) that compresses the raw tri-planes into a latent space (\cref{sec:triplane_fitting}). Latent codes are then generated by a diffusion model conditioned on partial observations, leveraging depth priors (\cref{sec:conditioning}). To further improve alignment with the input image, we apply depth-guided sampling during inference (\cref{sec:guided_sampling}).

\subsubsection{Tri-plane Fitting and VAE Training}\label{sec:triplane_fitting}

We represent 3D shapes using signed distance fields (SDFs), which are decoded from triplane features via a multilayer perceptron (MLP) parameterized by $\theta$. The tri-plane $\mathcal{F}\in \mathbb{R}^{3\times C_{\mathrm{raw}} \times N_{\mathrm{raw}}^2}$ is a tensor built on three axis-aligned 2D planes: XY, YZ, and XZ, to factorize the dense 3D volume grid. Here, $N_\mathrm{raw}$ denotes the spatial resolution and $C_\mathrm{raw}$ is the number of feature channels in the raw tri-plane. Given a query point $q \in \mathbb{R}^{3}$, the function $\Phi$: $\mathbb{R}^{3}\mapsto \mathbb{R}$ outputs its signed distance value:
\begin{equation}
\Phi_{\theta}(q)=\operatorname{MLP}_{\theta} 
\left( \underset{\Pi \in {\operatorname{XY,YZ,XZ}}}{\bigoplus} \mathcal{F}_{\Pi} \Bigl( \operatorname{Proj}_{\Pi}(q) \Bigr)  \right) \label{eqn:triplane}
\end{equation}
where $\operatorname{Proj}_{\Pi}(\cdot)$ orthogonally projects the 3D point onto plane $\Pi$ to retrieve plane features via bilinear interpolation, and $\oplus$ denotes feature concatenation.

To convert all meshes in the dataset into tri-plane representations, we follow the procedure in BlockFusion~\cite{wu2024blockfusion}. For each object, we first convert the geometry into a watertight mesh, then uniformly sample on-surface points $\Omega_0$ and off-surface points $\Omega$. We jointly optimize the MLP parameters $\theta$ and the tri-plane features $\mathcal{F}$ using the following geometry loss:
\begin{equation}
  \mathcal{L}_\mathrm{geometry}= \mathcal{L}_\mathrm{normal}+ \mathcal{L}_\mathrm{eikonal}+ \mathcal{L}_\mathrm{SDF}.
\end{equation}
where each term is defined as:
\begin{equation}
  \mathcal{L}_\mathrm{normal}= \sum_{p \in \Omega_0}\textbf{n}_p \cdot \hat{\textbf{n}}_p+ \left\|\textbf{n}_p-\hat{\textbf{n}}_{p}\right\|_{2}
\end{equation}
\begin{equation}
  \mathcal{L}_\mathrm{eikonal}= \sum_{p \in \Omega} \left\|\hat{\textbf{n}}_p-1\right\|_2
\end{equation}
\begin{equation}
  \mathcal{L}_\mathrm{SDF}= \sum_{p\in \Omega \cup \Omega_0}\left\| \Phi_{\theta}(p) - s_{p}\right\|
\end{equation}
where $\hat{\textbf{n}}_p=\nabla\Phi_{\theta}(p)$ is the predicted surface normal, $\textbf{n}_p$ is the ground truth surface normal, and $s_p$ is the ground truth signed distance value at point $p$. For points on the surface ($p \in \Omega_0$), $s_p = 0$. For brevity, we omit the weighting coefficients of individual loss terms.

To compress raw tri-planes into a lower-dimensional latent space, we employ a variational autoencoder (VAE). The encoder $\mathcal{E}$ maps $\mathcal{F}$ to a latent code $z \in \mathbb{R}^{3 \times C_\mathrm{latent} \times N_\mathrm{latent}^2}$, where $C_\mathrm{latent} < C_{\mathrm{raw}}$ and $N_\mathrm{latent} < N_{\mathrm{raw}}$. The decoder $\mathcal{D}$ reconstructs the raw tri-planes from the latent representation. The VAE is trained to minimize the following objective:
\begin{equation}
  \mathcal{L}_{\mathrm{VAE}}= \mathcal{L}_{\mathrm{recon}}+ \mathcal{L}_{\mathrm{KL}}+ \mathcal{L}_{\mathrm{geometry}}
\end{equation}
where $\mathcal{L}_{\mathrm{recon}}$ is the L1 loss between $\mathcal{F}$ and $\mathcal{D}
(\mathcal{E}(\mathcal{F}))$, $\mathcal{L}_\mathrm{KL}$ denotes KL divergence loss, and $\mathcal{L}_{\mathrm{geometry}}$ is computed
using the decoded latent.

\subsubsection{Depth-conditioned Denoising}\label{sec:conditioning}

\begin{figure}[ht]
    \centering
    \includegraphics[width=\linewidth]{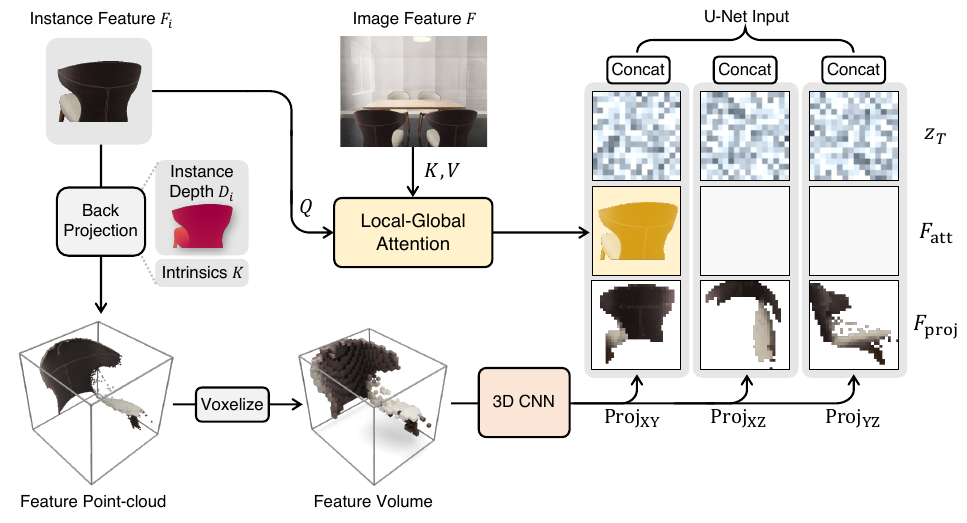}
    \caption{Illustration of depth-guided conditioning. \texttt{Concat} denotes channel-wise concatenation. The attention-enhanced feature $F_\mathrm{att}$ is applied to the XY plane, while zero padding is used for the remaining two planes.}
    \label{fig:conditioning}
\end{figure} 

After obtaining latent tri-plane representations for all objects, we design a diffusion model that generates latents for complete shapes, conditioned on partial observations. Rather than injecting image features through naive concatenation or cross-attention, we leverage strong geometric priors provided by depth.

Specifically, we back-project the instance features $F_i$ using the estimated depth $D_i$ and camera intrinsics $K$, and voxelize the resulting 3D points into a feature volume of size $N_\mathrm{latent}^3$, where the spatial resolution $N_\mathrm{latent}$ is chosen to match that of the latent tri-planes. Features falling into the same voxel are aggregated via average pooling. The voxelized feature volume $V$ is then processed by a 3D CNN and a projection layer: $V_\mathrm{conv} = \operatorname{Conv3D}(V)$.
Instead of reconstructing shapes directly from these features, as in previous works~\cite{zhang2023uni,dahnert2021panoptic}, we use them as conditioning inputs for our latent diffusion model. This generative formulation leverages shape priors from depth to improve shape fidelity, particularly in the presence of occlusion. After volumetric feature extraction, we project $V_\mathrm{conv}$ onto the three orthogonal planes (XY, YZ, XZ) using $\operatorname{Proj}_{\Pi}(\cdot)$ as defined in \cref{eqn:triplane}, followed by average pooling along the projection axis. The resulting 3-view feature maps $F_\mathrm{proj} = \left\{\operatorname{Proj}_{\Pi}(V_\mathrm{conv}) \;|\;\Pi \in \{ \mathrm{XY}, \mathrm{YZ}, \mathrm{XZ}\} \right\}$ provide a structured and spatially aligned representation of the partial observations, enriched with shape priors from the depth.

To incorporate global scene context, \eg, cues from other visible instances of the same category or shape, we introduce a local-global attention mechanism. It enhances the local object feature with global scene-level information. Given the instance features $F_i$ and the image features $F$, we compute cross-attention as:
\begin{equation}
F_{\mathrm{att}} = \operatorname{CrossAttention}(q=F_i, k=F, v=F)
\end{equation}

This local-global attention allows our model to incorporate broader visual context when reconstructing occluded or ambiguous regions.

The 3-view features $F_\mathrm{proj}$, together with the attention-enhanced features $F_\mathrm{att}$, are passed to a 3D-Aware U-Net~\cite{wu2024blockfusion} denoiser $\mathcal{G}$. These features are concatenated with the noised latent $z_t$ along the feature dimension and used as input to the network. An overview of the conditioning process is illustrated in \cref{fig:conditioning}. The model predicts the clean latent $z_0$ as
\begin{equation}
  \hat{z}_0 = \mathcal{G} \left(\operatorname{Concat}(z_t, F_\mathrm{proj}, F_\mathrm{att}), t\right)
\end{equation}
We train the denoiser using the standard diffusion objective:
\begin{equation}
\mathcal{L}_\mathrm{diffusion} = \mathbb{E}_{z_0, \epsilon, t} \left[ \|z_0 - \hat{z}_0\|_2^2 \right]
\end{equation}
where $z_t = \sqrt{\bar{\alpha}_t} z_0 + \sqrt{1 - \bar{\alpha}_t} \epsilon$, with $\epsilon \sim \mathcal{N}(\mathbf{0}, \mathbf{I})$ and $\bar{\alpha}_t$ denoting the noise schedule. To further preserve geometric fidelity, we incorporate the geometry loss $\mathcal{L}_\mathrm{geometry}$ from \cref{sec:triplane_fitting}. The final training objective is
\begin{equation}
\mathcal{L}_\mathrm{recon} = \mathcal{L}_\mathrm{diffusion} + \mathcal{L}_\mathrm{geometry}
\end{equation}

This conditioning framework enables our model to effectively integrate partial 3D observations with image-based contextual cues, producing accurate and high-fidelity 3D shapes. By combining depth guidance with local-global attention, our method remains robust to occlusions and ambiguities, resulting in consistent and detailed reconstructions.

\subsubsection{Depth-guided Sampling}\label{sec:guided_sampling}

Generative models may produce inaccurate shapes due to insufficient observations. However, depth estimators provide valuable priors for object shapes. To incorporate this information during inference, we adopt classifier-guided DDIM~\cite{song2020denoising} sampling, using depth to guide SDF generation. Specifically, we render the predicted sample $\hat{z}_0$ obtained during the diffusion process into a depth map $\hat{D}_i$ using the differentiable SDF depth renderer from MonoSDF~\cite{yu2022monosdf}. We then compute the depth loss $\mathcal{L}_{\mathrm{depth}}$ against the estimated instance depth $D_i$, and use the gradient $\nabla_{\hat{z}_0}\mathcal{L}_{\mathrm{depth}}$ to guide the sampling trajectory.

To handle depth scale ambiguity in depth supervision, we employ scale-invariant loss \cite{eigen2014depth}, defined as
\begin{equation}
\resizebox{0.88\linewidth}{!}{
$
\mathcal{L}_{\mathrm{depth}} = \frac{1}{m} \sum\limits_{j=1}^{m} \left\| \left( \log \hat{D}_i^{(j)} - \log D_i^{(j)} + \alpha(\hat{D_i}, D_i)\right) \right\|_2
$
}
\end{equation}
where
\begin{equation}
  \alpha(\hat{D_i}, D) = \frac{1}{m} \sum_{j=1}^m \left( \log D_i^{(j)} - \log \hat{D}_i^{(j)} \right)
\end{equation}
$m$ is the number of valid pixels for instance depth $D_i$.

During diffusion sampling, we apply depth guidance to improve the geometric accuracy of the generated shape. At each timestep $t$, after obtaining the predicted clean latent $\hat{z}_0$, we compute the gradient of the depth loss and use it to update the predicted noise $\epsilon_t$ as:
\begin{equation}
\epsilon_t' = \epsilon_t - \lambda_{\mathrm{guide}} \sqrt{1-\bar{\alpha}_t} \nabla_{\hat{z}_0}\mathcal{L}_{\mathrm{depth}}
\end{equation}
where $\lambda_{\mathrm{guide}}$ is a guidance strength parameter that controls the influence of the depth prior, and $\sqrt{1-\bar{\alpha}_t}$ scales the gradient according to the current noise level. The adjusted noise $\epsilon_t'$ is then used in the DDIM update rule to compute the next latent $z_{t-1}$.

This depth-guided sampling process effectively steers generation toward geometry consistent with observed depth, leading to more accurate reconstruction, particularly for occluded objects.

To ensure proper alignment with the estimated depth map, we adopt a two-stage sampling strategy. In the first stage, we generate an initial 3D shape without applying depth guidance. Based on this initial output, we estimate the object pose using the layout estimator $F_\mathrm{layout}$, which will be described in \cref{sec:scene_comp}. Once the transformation parameters are obtained, we perform a second sampling pass with depth guidance. In this pass, the intermediate shape is transformed from canonical to camera space to render the depth map, and the scale-invariant loss is computed to guide further refinement.

\begin{table*}[t]
  \centering
  \resizebox{\linewidth}{!}{
  \begin{tabular}{lccccccc}
    \toprule
    \multirow{2}{*}[-2pt]{\textbf{Method}} & \multicolumn{3}{c}{\textbf{Scene-level}} & \multicolumn{2}{c}{\textbf{Object-level}} & \multirow{2}{*}[-2pt]{\textbf{Runtime}}\\
    \cmidrule(lr){2-4} \cmidrule(lr){5-6}
    & \textbf{CD ($\times 10^{-3}$, $\downarrow$)} & \textbf{CD-S ($\times 10^{-3}$, $\downarrow$)} & \textbf{F-Score (\%, $\uparrow$)} & \textbf{CD ($\times 10^{-3}$, $\downarrow$)} & \textbf{F-Score (\%, $\uparrow$)} \\
    \midrule
    InstPIFu~\cite{liu2022towards} & 213.4 & 124.9 & 13.72 & 44.74 & 29.63 & \qty{19.8}{\second}\\
    Uni-3D~\cite{zhang2023uni} & 218.3 & 113.3 & 12.99 & --- & --- & \qty{3.1}{\second} \\
    BUOL~\cite{chu2023buol} & 282.7 & 154.7 & 12.57 & --- & --- & \qty{6.5}{\second} \\
    Gen3DSR~\cite{dogaru2024generalizable} & 222.4 & 137.5 & 13.52 & 9.74 & 31.42 & \qty{15.5}{\minute} \\
    DeepPriorAssembly~\cite{zhou2024zero} & 191.8 & 76.2 & 16.72 & 20.13 & 27.83 & \qty{6.7}{\minute} \\
    \midrule
    \textbf{\OURS{} (Ours)} & \textbf{153.2} & \textbf{56.4} & \textbf{25.00} & \textbf{2.57} & \textbf{89.66} & \qty{1.2}{\minute} \\
    \bottomrule
  \end{tabular}
  }
  \caption{Quantitative comparison with state-of-the-art methods on the 3D-FRONT~\cite{fu20213d} test set. We report Chamfer distance (CD; lower is better) and F-score (higher is better) at scene and object levels, along with single-direction Chamfer distance (CD-S) at the scene level.}
  \label{tab:comparison}
\end{table*}

\subsection{Scene Composition}\label{sec:scene_comp}
As described in \cref{sec:obj_recon}, objects are reconstructed in their canonical coordinate systems, which enables flexibility in handling variations in pose and scale commonly found in indoor environments. Following DeepPriorAssembly~\cite{zhou2024zero}, we formulate scene composition as a shape registration problem, where each object’s transformation is parameterized by translation (3 parameters), rotation (1 parameter) around the vertical axis, and scale (1 parameter). To account for camera roll and pitch, we additionally apply a global rotation that is shared across all objects.

We randomly initialize these learnable transformation parameters, denoted as $\phi$. The sampled point cloud $P^\prime_{i}$ from the reconstructed surface $O_i$ of object $i$ is transformed into the camera coordinate system as $P_{i}=f_\phi(P^\prime_i)$, where $f_\phi$ is the parameterized transformation function. We optimize $\phi$ using gradient descent. The  optimization objective is defined as:
\begin{equation}
\mathcal{L}_{\mathrm{layout}}= \mathcal{L}_\mathrm{CD}(\tilde{P}_i, P_i) +
\mathcal{L}_\mathrm{CD}(\mathcal{P}(\tilde{P}_i), \mathcal{P}(P_i)) \label{eqn:layout}
\end{equation}
where $\tilde{P}_{i}$ is the partial point cloud from the depth map. $\mathcal{L}_\mathrm{CD}$ represents the Chamfer distance loss. The operator $\mathcal{P}$ denotes projection from 3D camera coordinates to 2D pixel space, where the Chamfer distance is computed between the corresponding 2D point sets.

We select the optimized parameters $\phi$ that yield the minimum 3D Chamfer distance (the first term in \cref{eqn:layout}) as the final object poses for scene reconstruction.

\begin{figure*}[ht]
  \centering
  \includegraphics[width=\linewidth,trim=0 1em 0 0,clip]{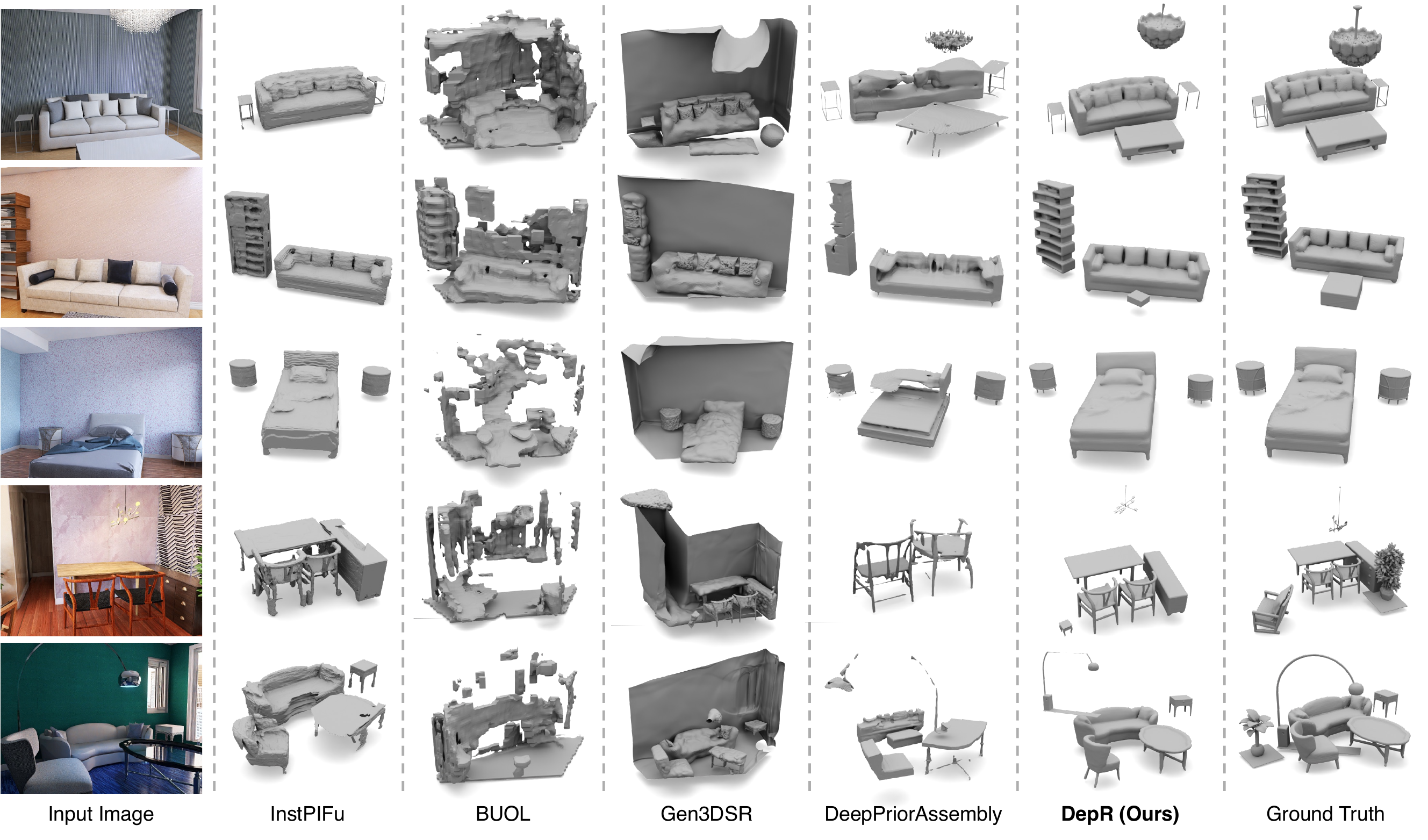}
  \caption{Qualitative comparisons the on the 3D-FRONT~\cite{fu20213d} synthetic dataset against state-of-the-art methods.}
  \label{fig:qualitative_synth}
\end{figure*}

\section{Experiments}
\label{sec:experiments}

\subsection{Datasets}
We conduct our experiments using the 3D-FRONT~\cite{fu20213d} dataset, specifically the version processed by InstPIFu~\cite{liu2022towards}. In addition to the original object meshes from the 3D-FUTURE~\cite{fu20213dfuture} dataset and scene layouts, it provides realistic rendered images and rich annotations such as depth maps, instance segmentation masks, and object labels. We use 22,673 scene images for training, and randomly sample 969 images for object-level evaluation and 156 images for scene-level evaluation.

\subsection{Implementation Details}
We use off-the-shelf models to ensure 
robustness and broad applicability: Grounded-SAM~\cite{ren2024grounded} for instance segmentation, Depth Pro~\cite{bochkovskii2024depth}
for depth estimation, and DINOv2~\cite{oquab2023dinov2} as the image encoder. Our raw tri-plane representation has a resolution of $128 \times 128$ with 32 feature channels, while the latent triplane is compressed to $32 \times 32$ with 2 channels. Prior to training, we perform a one-time tri-plane fitting across the entire 3D-FUTURE~\cite{fu20213dfuture} dataset, which contains approximately 16,000 shapes. This procedure is parallelized and takes around 24 hours on CPUs. The VAE and diffusion models are trained 
on four NVIDIA A6000 GPUs, for roughly 168 hours and 96 hours, respectively. During guided sampling, we set the gradient scale $s$ to 100.

\subsection{Metrics}

We evaluate our method using Chamfer distance (CD) and F-score (with a threshold of 0.002), following InstPIFu. Each reconstructed mesh is uniformly sampled with 10k points for metric computation.

We report metrics at two levels:

\begin{itemize}
 \item \textbf{Object-level:} To assess object quality, we decompose the predicted scene into individual instances and compare each against its corresponding ground truth mesh. For fair comparison, both predicted and ground-truth meshes are normalized to a unit sphere.

 \item \textbf{Scene-level:} We evaluate the full reconstructed scene, capturing both object geometry and their spatial layout. Following DeepPriorAssembly~\cite{zhou2024zero}, we additionally report the single-direction Chamfer distance (CD-S) from the generated scene to the ground truth.
\end{itemize}

\begin{figure*}[ht]
  \centering
  \includegraphics[width=\linewidth,trim=0 1em 0 0,clip]{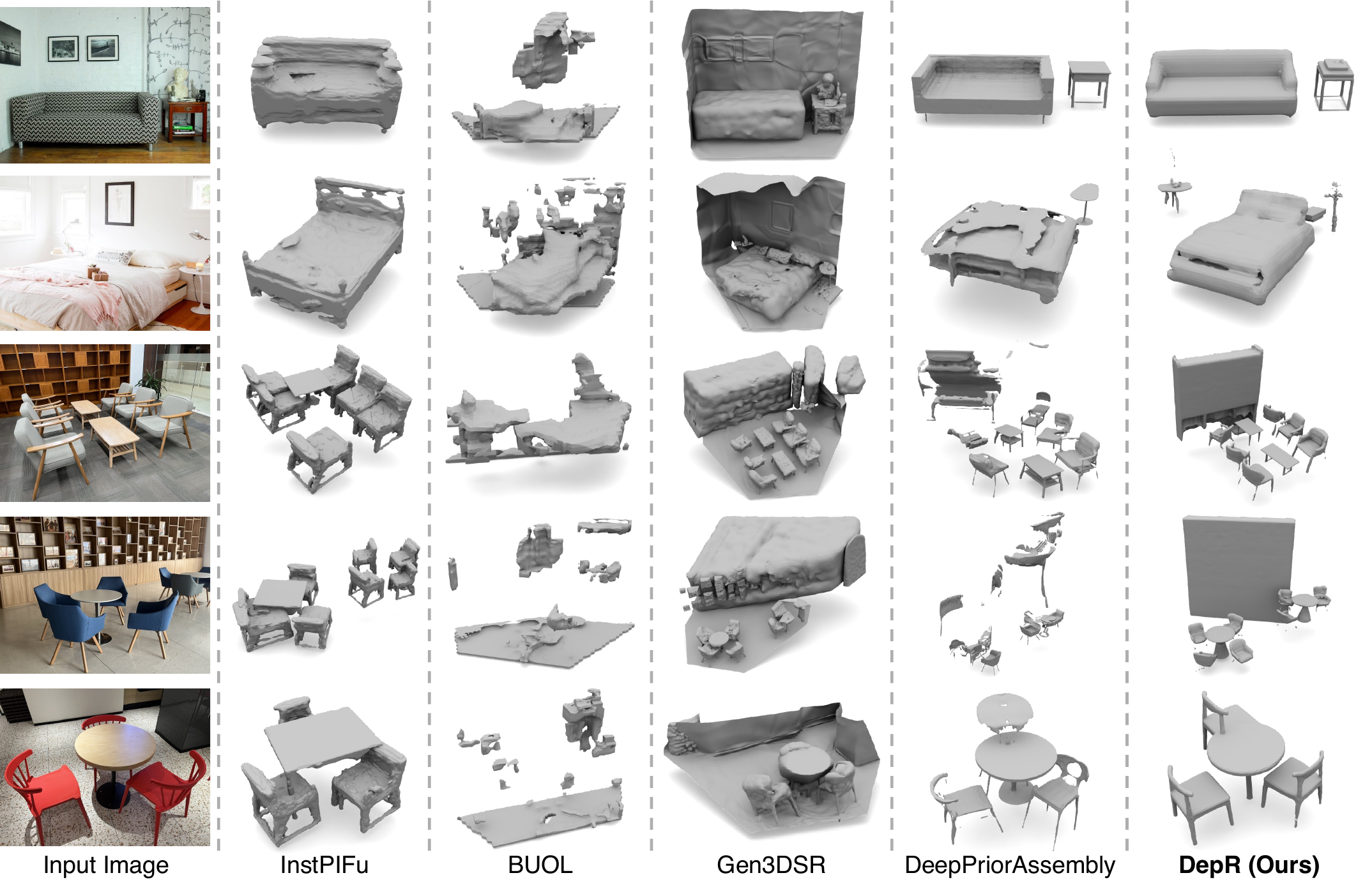}
  \caption{Qualitative comparisons on real images, including Pix3D~\cite{sun2018pix3d} (top two rows) and our own collected images (bottom three rows).}
  \label{fig:qualitative_real}
\end{figure*}

\subsection{Results}

We benchmark our method against two categories of approaches: (1) generative-model-based methods, including Gen3DSR~\cite{dogaru2024generalizable} and DeepPriorAssembly~\cite{zhou2024zero}; and (2) feed-forward reconstruction methods, such as InstPIFu, Uni-3D~\cite{zhang2023uni}, and BUOL~\cite{chu2023buol}. Since Uni-3D and BUOL perform holistic scene reconstruction without generating individual object meshes, object-level evaluation is not applicable for these methods.

\cref{tab:comparison} presents a quantitative comparison between our method and state-of-the-art approaches on the 3D-FRONT test set. \OURS{} outperforms all baselines across both object-level and scene-level metrics. Notably, at the object level, it achieves a substantial performance gain, highlighting the effectiveness of our depth-guided latent diffusion in generating high-fidelity geometries, even under occlusion. At the scene level, our method consistently surpasses existing approaches, showcasing its ability to more accurately recover spatial relationships and maintain global consistency. Furthermore, our method achieves significantly lower inference time per scene (measured on a single NVIDIA A100 GPU) compared to other generative-model-based compositional approaches, highlighting its efficiency alongside reconstruction quality.

We present qualitative comparisons in \cref{fig:qualitative_synth} on the synthetic 3D-FRONT dataset, and in \cref{fig:qualitative_real} on real-world images from Pix3D~\cite{sun2018pix3d} and our own collected samples. In both settings, \OURS{} yields more accurate and visually coherent reconstructions. Its depth-guided design enhances object geometries and spatial arrangements, resulting in outputs that closely align with the input images. Notably, our approach generalizes well to real-world scenarios, successfully reconstructing scenes from diverse environments and demonstrating strong robustness and practicality.

\subsection{Ablation Study}

\begin{figure}[htb]
    \centering
    \includegraphics[width=\linewidth,trim=0 1em 0 0,clip]{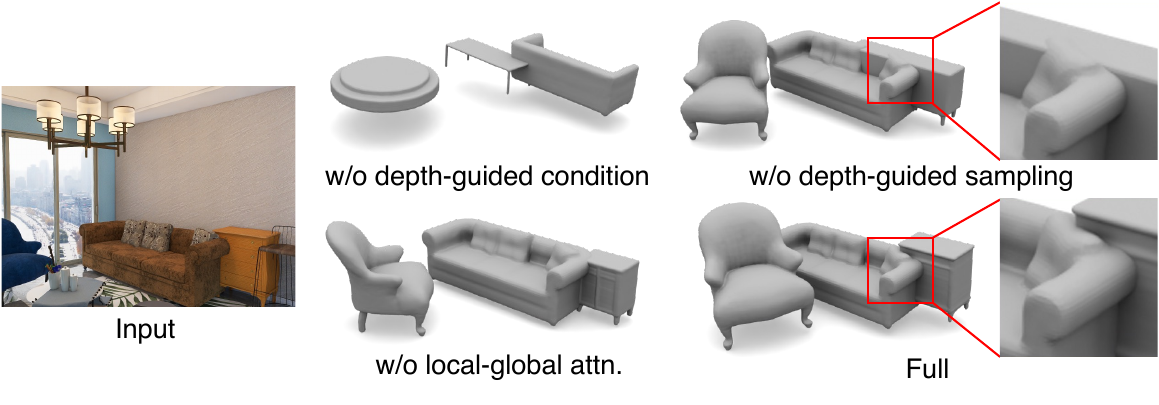}
    \caption{Qualitative comparisons on proposed components.}
    \label{fig:ablation_design}
    \vspace{2pt}
\end{figure}

\paragraph{Framework design.} To evaluate the efficacy of each proposed component, we conduct ablation studies and report object reconstruction results in \cref{tab:ablation_design}. Depth-guided conditioning yields the most significant improvement, boosting F-score by 27.76. Local-global attention provides a 2.3-point gain, while depth-guided sampling offers a smaller but noticeable enhancement. Qualitative results in \cref{fig:ablation_design} further illustrate that guided sampling improves alignment of fine details with the input image.

\begin{table}[htbp]
    \centering
    \resizebox{\linewidth}{!}{
    \setlength{\tabcolsep}{1em}
    \begin{tabular}{lccc}
    \toprule
    \multirow{2}{*}{\textbf{Method}} & \multicolumn{2}{c}{\textbf{Object-level}} \\
     & \textbf{CD ($\times 10^{-3}$, $\downarrow$)} & \textbf{F-Score ($\uparrow$)} \\
    \midrule
    Full model & \textbf{2.57} & \textbf{89.66} \\
    w/o Depth-guided Conditioning & 11.87 & 61.90 \\
    w/o Local-global Attention & 3.14 & 87.36 \\
    w/o Depth-guided Sampling & 2.61 & 89.64 \\
    \bottomrule
    \end{tabular}
    }
    \caption{Ablation studies on the proposed components.}
    \label{tab:ablation_design}
    \vspace{-0.5em}
\end{table}

\begin{table}[htbp]
    \centering
    \setlength{\tabcolsep}{1.5em}
    \resizebox{\linewidth}{!}{
    \begin{tabular}{cccc}
    \toprule
    \multirow{2}{*}{\textbf{GT Depth}} & \multirow{2}{*}{\textbf{GT Layout}} & \multicolumn{2}{c}{\textbf{Object-level}}  \\
     & & \textbf{CD ($\times 10^{-3}$, $\downarrow$)} & \textbf{F-Score ($\uparrow$)} \\
    \midrule
    \multicolumn{4}{c}{\cellcolor{gray!15}\textbf{w/o Depth-guided Sampling}} \\
               &  $-$ & 2.61 & 89.64 \\
    \checkmark &  $-$ & 2.42 & 90.33 \\
    \midrule
    \multicolumn{4}{c}{\cellcolor{gray!15}\textbf{with Depth-guided Sampling}} \\
                           &            & 2.57          & 89.66 \\
                      & \checkmark & 2.55          & 89.92 \\
    \checkmark  &            & 2.37          & 90.24 \\
    \checkmark  & \checkmark & \textbf{2.29} & \textbf{90.62} \\
    \bottomrule
    \end{tabular}
    }
    \caption{Effects of depth and layout in object reconstruction.}
    \label{tab:ablation_obj_recon}
    \vspace{-0.5em}
\end{table}

\vspace{-6pt}
\paragraph{Effects of depth and layout in object reconstruction.} Depth is used both for conditioning object reconstruction and for layout estimation during guided sampling. In \cref{tab:ablation_obj_recon}, we present ablations using ground-truth depth for conditioning and layout estimation, and additionally comparing with ground-truth layout to estimate the theoretical upper bound of model performance. Results show that ground-truth depth further improves object reconstruction quality. Moreover, guided sampling with ground-truth layout yields a more notable improvement (approximately $+$0.3 F-score) over the non-guided variant, regardless of whether ground-truth depth is used for conditioning. This suggests that improved layout estimation can further benefit the model’s performance.

\begin{table}[htbp]
    \centering
    \resizebox{\linewidth}{!}{
    \begin{tabular}{cccccc}
    \toprule
    \multicolumn{3}{c}{\textbf{GT Inputs}} & \multicolumn{3}{c}{\textbf{Scene-level}} \\
    \textbf{Depth} & \textbf{Segm} & \textbf{Layout} & \textbf{CD ($\times 10^{-3}$, $\downarrow$)} & \textbf{CD-S ($\times 10^{-3}$, $\downarrow$)} & \textbf{F-Score ($\uparrow$)} \\
    \midrule
                   &            &            & 155.0 & 57.3 & 24.63 \\
        \checkmark &            &            & 152.4 & 61.4 & 26.09 \\
                   & \checkmark &            & 103.4 & 34.1 & 30.47 \\
                   & \cellcolor{gray!15} \checkmark & \cellcolor{gray!15} \checkmark & \underline{20.2}  & \underline{10.5} & \textbf{67.65} \\
        \checkmark & \checkmark &            & 91.6  & 28.0 & 33.14 \\
        \checkmark & \cellcolor{gray!15} \checkmark & \cellcolor{gray!15} \checkmark & \textbf{18.8}  & \textbf{9.5}  & \underline{67.42} \\
    \bottomrule
    \end{tabular}
    }
    \caption{Effects of upstream (depth, segmentation, and layout) errors in scene reconstruction. Note that ground-truth layout implies ground-truth segmentation.}
    \label{tab:ablation_scene_recon}
    \vspace{-0.5em}
\end{table}

\vspace{-0.5em}
\paragraph{Effects of upstream errors in scene reconstruction.} \OURS{} relies on external models for depth estimation and segmentation, making it susceptible to error propagation. In \cref{tab:ablation_scene_recon}, we show ablations using ground-truth depth, segmentation, and layout to assess their individual impact on scene-level reconstruction. We disable depth-guided sampling to isolate the effect of layout from object reconstruction. Note that ground-truth layout implies ground-truth segmentation due to their one-to-one correspondence. Results show that while ground-truth depth improves layout estimation, segmentation quality has a more pronounced effect as missing or incorrectly segmented objects directly degrade scene composition. The main limitation stems from optimization-based layout estimation; with ground-truth layout, scene-level F-score exceeds 67.

\section{Conclusion} \label{sec:conclusion}
In this paper, we present \OURS{}, a depth-guided single-view scene reconstruction framework that integrates depth priors into both object reconstruction and scene layout estimation. By leveraging depth-guided conditioning, our model achieves higher shape fidelity and improved robustness to occlusions. A local-global attention mechanism further incorporates scene-level context to refine object reconstructions. Additionally, we introduce depth-guided DDIM sampling to enhance geometric consistency between the reconstructed 3D scene and the input image. Extensive quantitative and qualitative evaluations demonstrate that \OURS{} outperforms state-of-the-art methods at both the object and scene levels, while generalizing effectively to real-world images. Our approach highlights the critical role of depth priors in single-view reconstruction and lays the groundwork for more robust and scalable 3D scene understanding.

\section*{Acknowledgment} This work is supported by NSF award IIS-2127544 and NSF award IIS-2433768.
We thank Lambda, Inc. for their compute resource help and travel grant.

{
    \small
    \bibliographystyle{ieeenat_fullname}
    \bibliography{main}
}
\clearpage
\appendix
\counterwithin{figure}{section}
\counterwithin{table}{section}
\counterwithin{equation}{section}
\setcounter{table}{0}
\setcounter{figure}{0}
\setcounter{equation}{0}
\section{Appendix}

\subsection{Detailed Runtime Analysis}

\cref{tab:inference} reports the per-module inference time for \OURS{}, averaged per scene and measured on a single NVIDIA A100 GPU. The 2D preprocessing stage, which includes segmentation and depth estimation using pre-trained models, takes \qty{1.6}{\second} in total. Since diffusion is performed over latent tri-planes of size $2 \times 32^2$, as opposed to the raw tri-plane of size $32\times 128^2$, the process remains efficient, requiring only \qty{1.2}{\second} for 50 DDIM sampling steps.

The most computationally expensive operation is guided sampling, which involves depth map rendering at each sampling step and takes \qty{35.9}{\second}. Layout optimization requires \qty{16.1}{\second} and is performed twice during guided sampling.

Overall, the full pipeline takes approximately \qty{1.2}{\minute} per scene with guided sampling enabled, and about \qty{20}{\second} without it.

\begin{table}[ht]
    \centering
    \caption{Inference runtime of different modules for \OURS{}.}\label{tab:inference}
    \begin{tabular}{l c}
    \toprule
    \textbf{Module}  & \textbf{Runtime} \\
    \midrule
    Segmentation  & \qty{0.8}{\second}   \\
    Depth Estimation  & \qty{0.8}{\second} \\
    Latent Triplane Diffusion & \qty{1.2}{\second} \\
    VAE + SDF Decoding & \qty{1.0}{\second}    \\
    Layout Optimization & \qty{16.1}{\second}    \\
    Guided Sampling & \qty{35.9}{\second}    \\
    \bottomrule
    \end{tabular}
    \label{tab:runtime}
\end{table}

\subsection{Additional Implementation Details}
\subsubsection{SDF Depth Rendering}
Depth-guided sampling requires rendering a depth map for the object being reconstructed. Following MonoSDF~\cite{yu2022monosdf}, we render the predicted SDF field into a depth map via differentiable volumetric rendering. SDF values $s$ are first converted into density values using the following equation:
\begin{equation}
  \sigma_{\beta}(s) =
  \begin{cases}
    \frac{1}{2\beta} \exp\left(\frac{s}{\beta}\right)                              & \text{if } s \leq 0 \\
    \frac{1}{\beta} \left(1 - \frac{1}{2} \exp\left(-\frac{s}{\beta}\right)\right) & \text{if } s > 0
  \end{cases}
\end{equation}
where $\beta$ is a hyper-parameter, set to $0.001$ in our experiments.

To compute the depth $\hat{D}(\mathbf{r})$ of the surface intersecting the current ray $\mathbf{r}$, we sample $M$ points of the form $\mathbf{o} + t_\mathbf{r}^i \mathbf{v}$, where $\mathbf{o}$ and $\mathbf{v}$ are the camera origin and viewing direction, respectively. The expected depth is computed as:
\begin{equation}
  \hat{D}(\mathbf{r}) = \sum_{i=1}^{M}T_{\mathbf{r}}^{i}\alpha_{\mathbf{r}}^{i}t_{\mathbf{r}}^{i}
\end{equation}
where the transmittance $T_\mathbf{r}^{i}$ is defined as:
\begin{equation}
  T_\mathbf{r}^{i}= \prod_{j=1}^{i-1}(1 - \alpha_\mathbf{r}^{j})
\end{equation}
and the alpha value $\alpha_\mathbf{r}^{i}$ is given by:
\begin{equation}
    \alpha_\mathbf{r}^{i}= 1 - \exp(-\sigma_\mathbf{r}^{i}\delta_\mathbf{r}^{i})
\end{equation}
where $\delta_\mathbf{r}^{i}$ is the distance between adjacent sample points. The rendered depth $\hat{D}$ is then used to compute the scale-invariant loss, which guides the DDIM sampling process for improved alignment with the actual depth.

\subsubsection{Network Architecture}
For the conditioning input, we construct the feature volume $V$ by back-projecting image features into 3D space, followed by a 3D CNN and a linear projection to obtain a tensor of shape $9 \times 32^3$. After orthogonal projection onto the three planes (XY, YZ, XZ), we obtain 3-view feature maps $F_\mathrm{proj} \in \mathbb{R}^{3\times 9 \times 32^2}$.

We then concatenate the noised latent $z_t \in \mathbb{R}^{3\times 2\times 32^2}$, the 3-view features $F_\mathrm{proj}$, and the attention-enhanced 2D feature $F_\mathrm{att} \in \mathbb{R}^{1 \times 32^2}$ (zero-padded for XZ and YZ planes) along the feature dimension. The resulting tensor has shape $3\times 12 \times 32^2$ and is passed to the diffusion U-Net.

The diffusion U-Net follows the architecture proposed in BlockFusion~\cite{wu2024blockfusion}, which adapts a standard 2D U-Net to operate over tri-plane inputs. All 2D convolution layers are replaced with a specialized \texttt{GroupConv} operator, enabling parallel processing of the three planes. To allow information exchange across planes, the feature maps are flattened into 1D tokens and passed through six self-attention layers in the U-Net’s middle block.
 
\subsection{Additional Qualitative Results}

\begin{figure*}[ht]
  \centering
  \includegraphics[width=\linewidth]{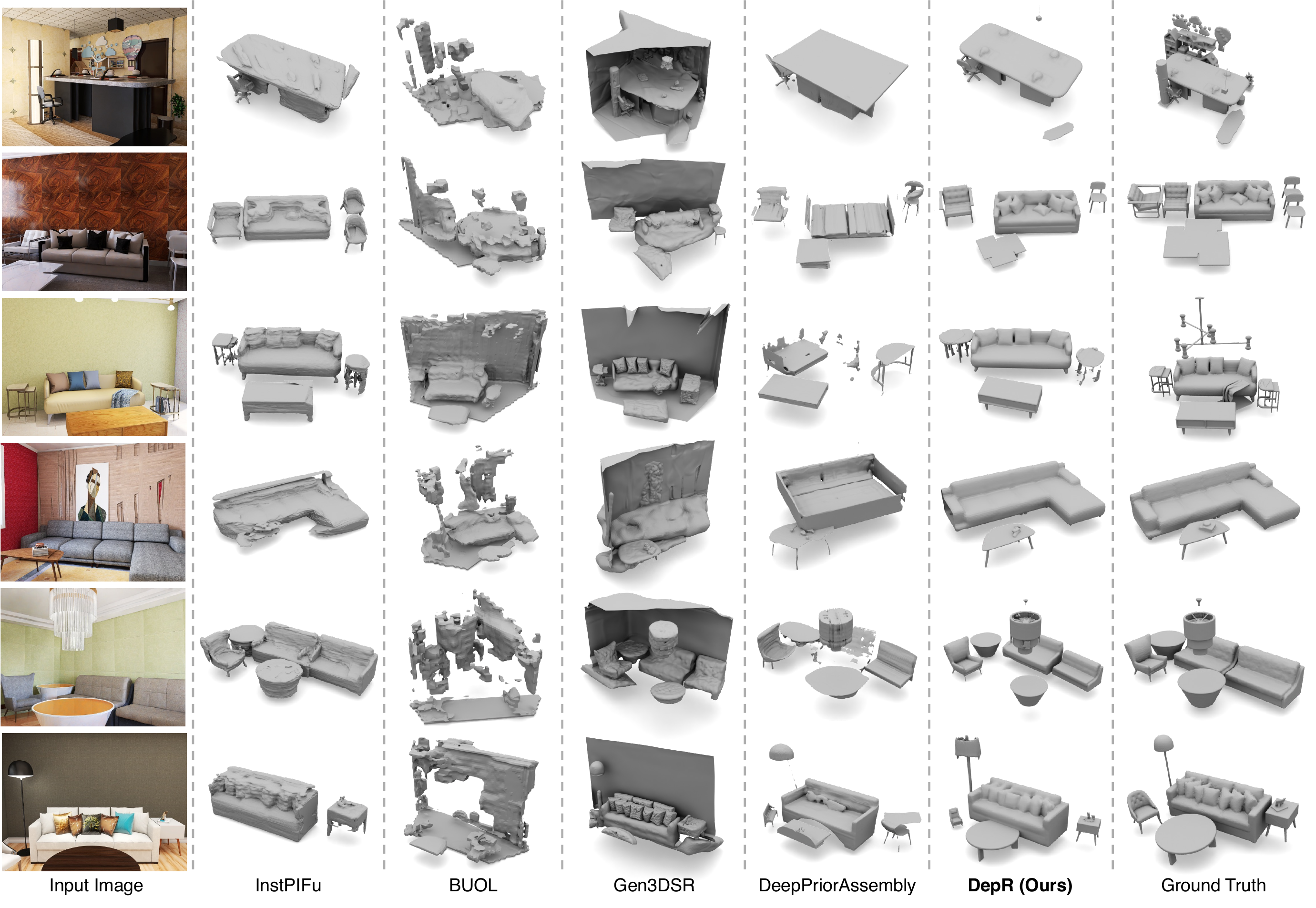}
  \caption{Additional qualitative results on the 3D-FRONT~\cite{fu20213d} dataset.}
  \label{fig:qualitative_additional}
\end{figure*}

\cref{fig:qualitative_additional} presents additional qualitative examples from the 3D-FRONT~\cite{fu20213d} dataset. Feed-forward methods such as InstPIFu~\cite{liu2022towards} and BUOL~\cite{chu2023buol} generally demonstrate limited generalizability. BUOL, which is trained on less realistic renderings from 3D-FRONT, fails to reconstruct meaningful geometry in most scenes, except for the scene in row 3. Among compositional methods, \OURS{} consistently produces more visually coherent surfaces and superior overall geometry.

\subsection{Limitations}

\begin{figure}[ht]
    \centering
    \includegraphics[width=\linewidth]{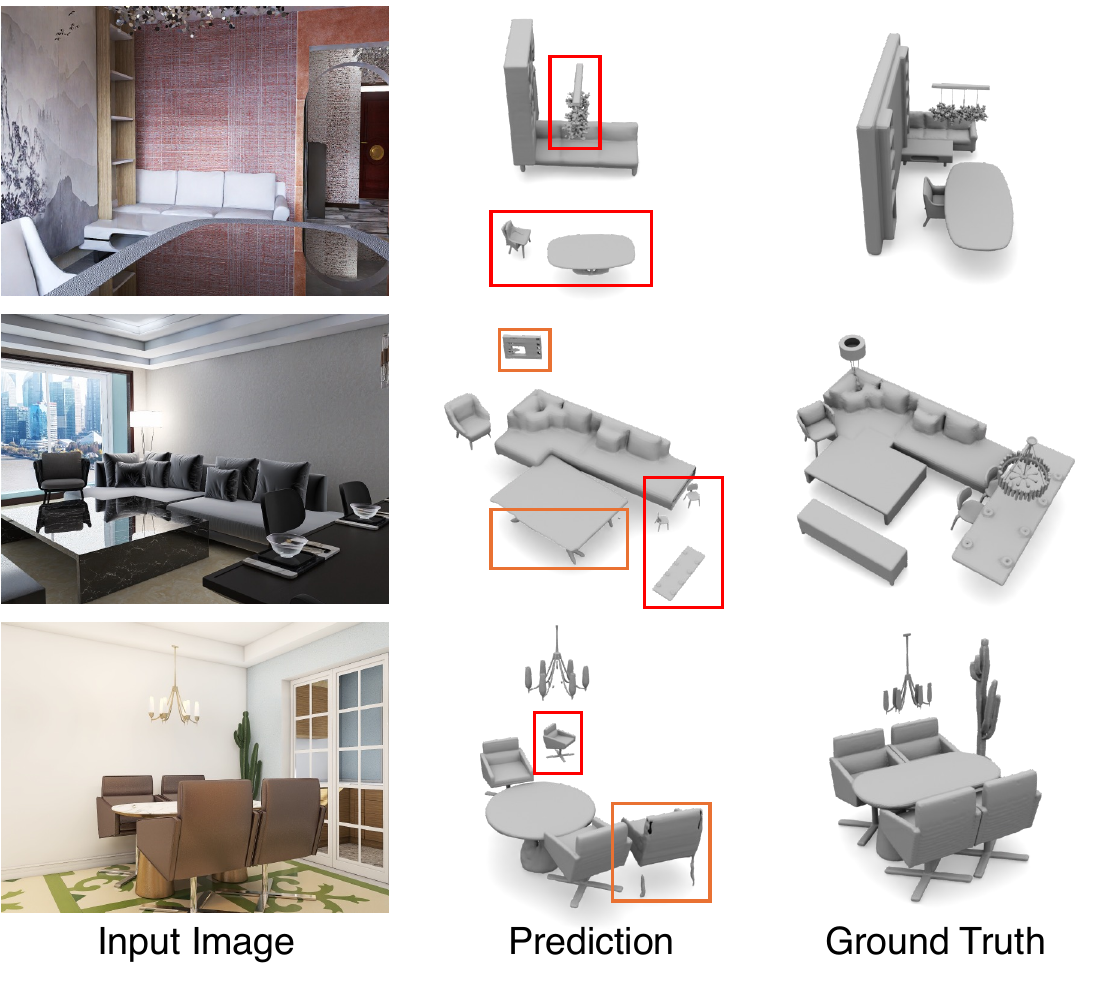}
    \caption{Failure cases of \OURS{}. Red rectangles indicate incorrect layout estimation; orange rectangles highlight incorrect object reconstruction.}
    \label{fig:failure_cases}
\end{figure}

\cref{fig:failure_cases} illustrates several failure cases of \OURS{}. Due to its generative nature, \OURS{} may incorrectly reconstruct or ``hallucinate'' objects when observations are severely limited by extreme occlusions (highlighted in orange rectangles). In the scene composition stage, the primary limitation arises from the optimization-based layout estimation (highlighted in red rectangles), which is susceptible to local minima. This issue becomes particularly pronounced when only a small portion of an object is visible, leading to highly incomplete depth point clouds that hinder accurate pose estimation.

Future work could address this limitation by integrating a learned, feed-forward pose regression module into the reconstruction framework. Such an approach may reduce the sensitivity to local minima and significantly improve overall inference efficiency.

\end{document}